\documentclass{article}

\usepackage{arxiv}

\usepackage[utf8]{inputenc} 
\usepackage[T1]{fontenc}    
\usepackage{hyperref}       
\usepackage{url}            
\usepackage{booktabs}       
\usepackage{amsfonts}       
\usepackage{nicefrac}       
\usepackage{microtype}      
\usepackage{lipsum}		
\usepackage{graphicx}
\usepackage{natbib}
\usepackage{doi}

\usepackage{subfigure}
\usepackage{verbatim}
\usepackage{amsmath,amssymb,amsfonts,pifont}
\usepackage{multirow}
\usepackage{algorithm}
\usepackage{algorithmic}
\usepackage{enumitem}

\DeclareMathOperator*{\argmin}{arg\,min}
\usepackage{bm}

\usepackage{hyperref}
\usepackage{pifont}%

\usepackage{todonotes}
\usepackage{multirow}

\usepackage{amsmath}
\usepackage{amssymb}
\usepackage{mathtools}
\usepackage{amsthm}

\theoremstyle{plain}

\theoremstyle{definition}

\theoremstyle{remark}

\title{\sysname: Few-shot Task-agnostic Neural Architecture Search for Distilling Large Language Models}

\author{Dongkuan Xu \\
	The Pennsylvania State University\\
	\texttt{dux19@psu.edu} \\
	\And
	Subhabrata Mukherjee \\
	Microsoft Research\\
	\texttt{submukhe@microsoft.com} \\
	\And
	Xiaodong Liu \\
	Microsoft Research\\
	\texttt{xiaodl@microsoft.com} \\
	\And
	Debadeepta Dey \\
	Microsoft Research\\
	\texttt{dedey@microsoft.com} \\
	\And
	Wenhui Wang \\
	Microsoft Research\\
	\texttt{wenwan@microsoft.com} \\
	\And
	Xiang Zhang \\
	The Pennsylvania State University\\
	\texttt{xzz89@psu.edu} \\
	\And
	Ahmed Hassan Awadallah \\
	Microsoft Research\\
	\texttt{hassanam@microsoft.com} \\
	\And
	Jianfeng Gao \\
	Microsoft Research\\
	\texttt{jfgao@microsoft.com} \\
}

\renewcommand{\headeright}{}
\renewcommand{\undertitle}{}
\renewcommand{\shorttitle}{Few-shot Task-agnostic Neural Architecture Search}

\begin{document}

\newcommand{\sysname}{\texttt{AutoDistil} }
\newcommand{\sysnameOnlyAuto}{{\footnotesize{ \texttt{AutoDistil}}}}
\newcommand{\cmark}{\ding{51}}%
\newcommand{\xmark}{\ding{55}}%

\maketitle

\begin{abstract}
	Knowledge distillation (KD) methods compress large models into smaller students with manually-designed student architectures given pre-specified computational cost. This requires several trials to find a viable student, and further repeating the process for each student or computational budget change.  We use Neural Architecture Search (NAS) to {automatically distill} several compressed students with variable cost from a large model. Current works train a single SuperLM consisting of millions of subnetworks with weight-sharing, resulting in interference between subnetworks of different sizes. Our framework \sysname addresses above challenges with the following steps: (a) Incorporates inductive bias and heuristics to partition Transformer search space into $K$ compact sub-spaces ($K$=$3$ for typical student sizes of base, small and tiny); (b) Trains one SuperLM for each sub-space using task-agnostic objective (e.g., self-attention distillation) with weight-sharing of students; (c) Lightweight search for the optimal student without re-training. Fully task-agnostic training and search allow students to be reused for fine-tuning on any downstream task. Experiments on GLUE benchmark against state-of-the-art KD and NAS methods demonstrate \sysname to outperform leading compression techniques with upto $2.7$x reduction in computational cost and negligible loss in task performance.
\end{abstract}

\section{Introduction}

\begin{figure}[!t]
\begin{center}
\centerline{\includegraphics[width=0.48\textwidth,height=0.32\textwidth]{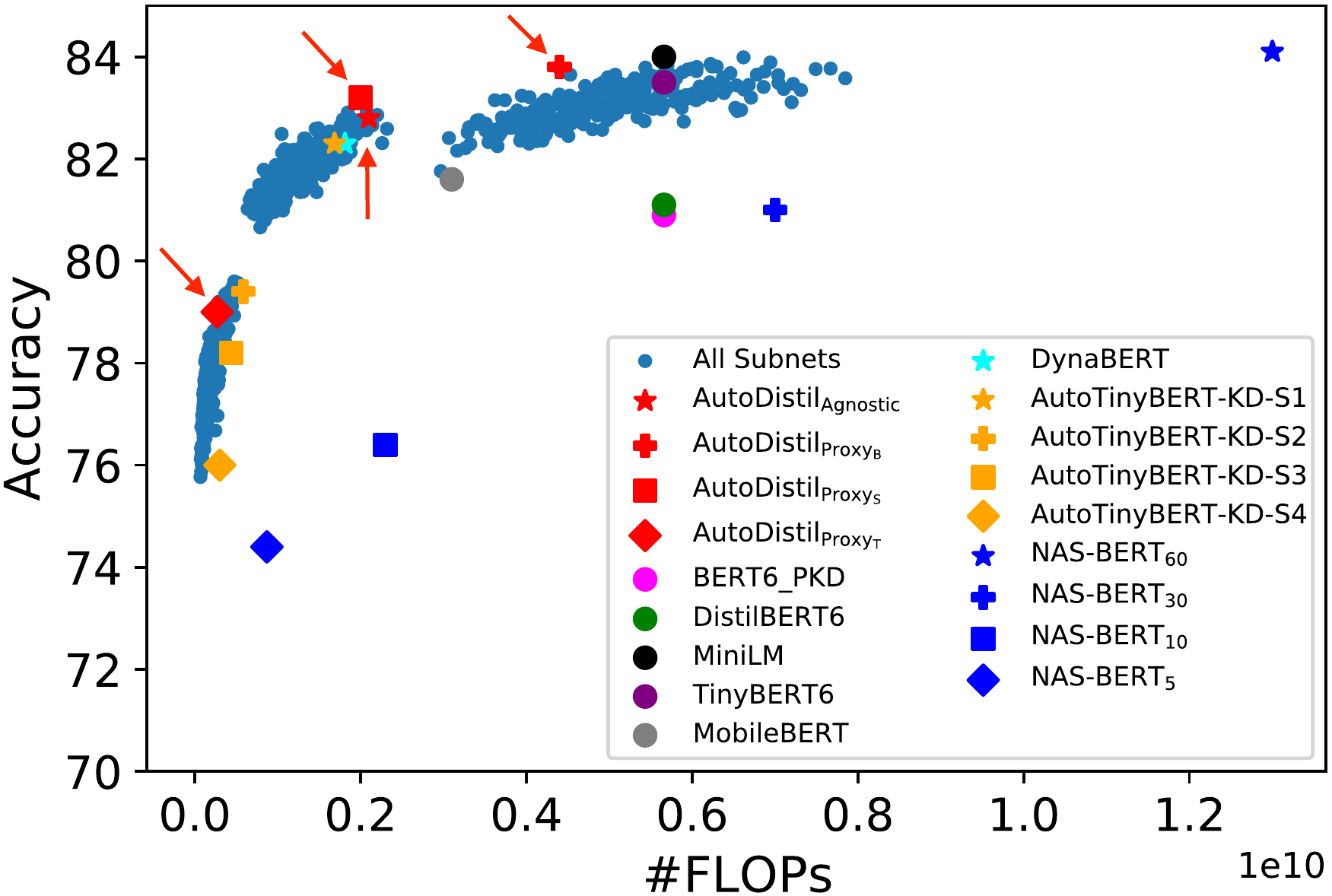}}
\vspace{0em}
\caption{\sysname uses few-shot task-agnostic Neural Architecture Search to distill several compressed students with variable \#FLOPs (x-axis) from $K$=$3$ SuperLMs (corresponding to each point cloud) trained on $K$ sub-spaces of Transformer search space. Each student (blue dot) extracted from the SuperLM is fine-tuned on MNLI with accuracy on y-axis. The best student from each SuperLM is marked in red. Given any state-of-the-art distilled model, \sysname generates a better candidate with less \#FLOPs and improved task performance from corresponding search space.}
\label{fig:mndr}
\end{center}
\vspace{-1.5em}
\end{figure}

While large pre-trained language models (e.g., BERT~\cite{devlin2019bert}, GPT-3~\cite{NEURIPS2020_1457c0d6}) are effective,  
their huge size poses significant challenges for downstream applications in terms of energy consumption and cost of inference~\cite{strubell2019energy} limiting their usage in on the edge scenarios and under constrained computational inference budgets.
%
Knowledge distillation~\cite{NEURIPS2020_3f5ee243,sanh2019distilbert,jiao2020tinybert,sun2020mobilebert}
has shown strong results in compressing pre-trained language models, where we train a small student model to mimic the full output distribution of the large teacher model. However, these works require pre-specification of the student model architecture and corresponding computational cost (e.g., number of parameters, FLOPs) before they can perform distillation. 
This poses two significant challenges: (i) since the architectures are hand-engineered, it requires several trials to come up with viable architectures and to define a myriad of hyper-parameters (e.g., number of layers, hidden dimension, number of attention heads, etc.); (ii) one has to re-run the distillation process with any change in specification for either the student architecture or the desired computational cost for using the student in a target environment. 

To address these challenges, Neural Architecture Search (NAS)~\cite{pham2018efficient,tan2019mnasnet,cai2020once,yu2020bignas} provides a natural solution to automatically search through a large space of candidate models while accounting for often conflicting objectives like computational cost vs. task performance. The dominant paradigm for NAS comprises of two main components: (a) Super model training which combines all possible architectures into a single graph and jointly training them via weight-sharing; and (b) Searching for optimal architecture from Super model with the best possible accuracy on a downstream task, satisfying a user-specified latency constraint for a specific device. 

NAS has demonstrated promising results in some recent explorations~\cite{NEURIPS2020_6f5216f8,yin-etal-2021-autotinybert,DBLP:conf/kdd/Xu0LS0QL21} 
in the natural language understanding domain. However, these works suffer from the following drawbacks. {\bf (D1)} All of these works train one single large Super Language Model (SuperLM) consisting of millions of diverse student architectures. This results in some undesirable effects of co-adaptation~\cite{bender2018understanding} like conflicts in weight-sharing where bigger student models converge faster in contrast to the smaller ones converging slower~\cite{zhao2021few,yu2020bignas}. Also, a single SuperLM may not have sufficient capacity to encode a large search space. As a result, these works use a multi-stage training process, where they first conduct NAS to identify candidate student models and then perform further pre-training~\cite{yin-etal-2021-autotinybert} and knowledge distillation~\cite{DBLP:conf/kdd/Xu0LS0QL21} of the candidates. {\bf (D2)} Additionally, these works are not fully task-agnostic. For instance, \citet{yin-etal-2021-autotinybert} performs task-agnostic SuperLM training, but task-specific search for the student with proxy tasks like SQuAD and MNLI. Similarly, \citet{DBLP:conf/kdd/Xu0LS0QL21} performs two-stage knowledge distillation with pre-training and fine-tuning of the candidates. Table~\ref{tb:baseline} contrasts \sysname with existing KD and NAS works.

We address these challenges with few-shot task-agnostic NAS consisting of the following three steps.

\noindent{\bf (S1) Search space design.} We partition the Transformer search space into $K$ sub-spaces ($K=3$ in our work for typical student model sizes like base, small and tiny) considering important architectural hyper-parameters like the network depth, width and number of attention heads. We further leverage inductive bias and heuristics to limit the number of student architectures in each sub-space.

\noindent{\bf (S2) Task-agnostic SuperLM training.} We train $K$ SuperLM, one for every sub-space. This allows each SuperLM more capacity to encode a sub-space as opposed to a single large one. We train each SuperLM with a task-agnostic objective like deep self-attention distillation, where we transfer knowledge from the self-attention module (including keys, queries and values) of a pre-trained teacher (e.g., BERT) to the student and use weight-sharing to train the SuperLM.

\noindent{\bf (S3) Lightweight optimal student search.} We obtain optimal student(s) directly from well-trained SuperLM(s) without any re-training. We propose two strategies to find the optimal student with task-agnostic or task-proxy search.

\begin{table}[!t]
\footnotesize
\addtolength{\tabcolsep}{-5.0pt}
\renewcommand{\arraystretch}{1.2}
\centering
\caption{Comparing \sysname with existing KD and NAS methods on aspects as task-agnostic training and search; generating multiple students with variable compression cost; single-stage training without additional adaptation; SuperLM training with compact search space to mitigate interference ($P$ denotes partial). }
\label{tb:baseline}
\begin{tabular}{lcc|ccc}
\toprule
{Method} & {Task-} & Variable  &  \multicolumn{3}{c}{NAS} \\
& agnostic & Compression & {Single}   & SuperLM    & Compact \\
 &  &   & {Stage}   & Training    & Search \\
\midrule
BERT-PKD  & \xmark & \xmark & \multicolumn{3}{c}{\multirow{6}{*}{N/A}} \\
SparseBERT & \xmark &  \xmark \\
DistilBERT & \checkmark & \xmark \\
TinyBERT  & \checkmark & \xmark  \\
MOBILEBERT & \checkmark & \xmark  \\
MINILM     & \checkmark & \xmark  \\\midrule
DynaBERT   & \xmark & \checkmark & \checkmark & One-shot  & \xmark    \\
NAS-BERT   & $P$ & \checkmark & \xmark & One-shot  & \xmark \\
AutoTinyBERT & $P$ & \checkmark & \xmark & One-shot & \xmark\\
\sysname      & \checkmark & \checkmark  & \checkmark & Few-shot  & \checkmark  \\
\bottomrule
\end{tabular}
\vspace{0em}
\end{table}

Overall, our contributions can be summarized as:

\noindent{\bf (1)} We develop a few-shot task-agnostic Neural Architecture Search framework to distill {several compressed models with variable computational cost}. We address the challenge of co-adaptation and weight-sharing of compressed models by few-shot NAS and a compact search space design.

\noindent{\bf (2)} We use self-attention distillation to train the SuperLM and demonstrate this to be better than masked language modeling objective for task-agnostic SuperLM training.

\noindent{\bf (3)} We perform extensive experiments in the GLUE benchmark where our method achieves $62.4\%$ reduction in computational cost and $59.7\%$ reduction in model size over state-of-the-art task-agnostic distillation methods with similar downstream task performance, with a comprehensive summary of the results in Figure~\ref{fig:mndr}.

\section{Background}
\label{sec:background}
We present an overview of Transformers~\cite{vaswani2017attention}, especially its two main sub-layers, multi-head self-attention (MHA) and feed-forward network (FFN).
Transformer layers are stacked to encode contextual information for input tokens as:  
\begin{align}
\label{eq:stacked_trans}
\mathbf{X}^l & = {\rm{Transformer}}_{l}(\mathbf{X}^{l-1}), \ l \in [1, L]
\end{align}
where $L$ is the number of Transformer layers, $\mathbf{X}^l$ $\in$ $\mathbb{R}^{s*d_{hid}}$, $s$ is the sentence length, and $d_{hid}$ is the hidden dimension. In the following, we omit the layer indices for simplicity.

\noindent {\bf Multi-Head Self-Attention (MHA).} Given the previous Transformer layer's output $\mathbf{X}$, the MHA output is given as:
\begin{align}
\vspace{-0em}
\label{eq:mha}
\mathbf{Q}_h, \mathbf{K}_h, \mathbf{V}_h  = \mathbf{X}&\bm{W}^Q_h, \mathbf{X}\bm{W}^K_h, \mathbf{X}\bm{W}^V_h, \\
{\rm{Attention}}(\mathbf{Q}_h, \mathbf{K}_h, \mathbf{V}_h) & = {\rm{softmax}}(\frac{\mathbf{Q}_h\mathbf{K}_h^{\top}}{\sqrt{d_{head}}})\mathbf{V}_h, \\
{\rm{MHA}}(\mathbf{X})  = {\rm{Concat}}&({\rm{head}}_1, \cdots, {\rm{head}}_H)\bm{W}^O,
\vspace{-0.5em}
\end{align}
where $\bm{W}^Q_h$, $\bm{W}^K_h$, $\bm{W}^V_h$ $\in$ $\mathbb{R}^{d_{hid}*d_{head}}$, $\bm{W}^O$ $\in$ $\mathbb{R}^{d_{hid}*d_{hid}}$ are linear transformations. $\mathbf{Q}_h$, $\mathbf{K}_h$, $\mathbf{V}_h$ $\in$ $\mathbb{R}^{s*d_{head}}$ are called queries, keys, and values, respectively. $H$ is the number of heads. ${\rm{head}}_h$ $=$ ${\rm{Attention}}(\mathbf{Q}_h, \mathbf{K}_h, \mathbf{V}_h)$ denotes the $h$-th attention head. ${\rm{Concat}}$ is the concatenating operation. $d_{head}$ $=$ $d_{hid}/H$ is the dimension of each head.

\begin{figure*}[!t]
\begin{center}
\centerline{\includegraphics[width=0.95\textwidth,height=0.425\textwidth]{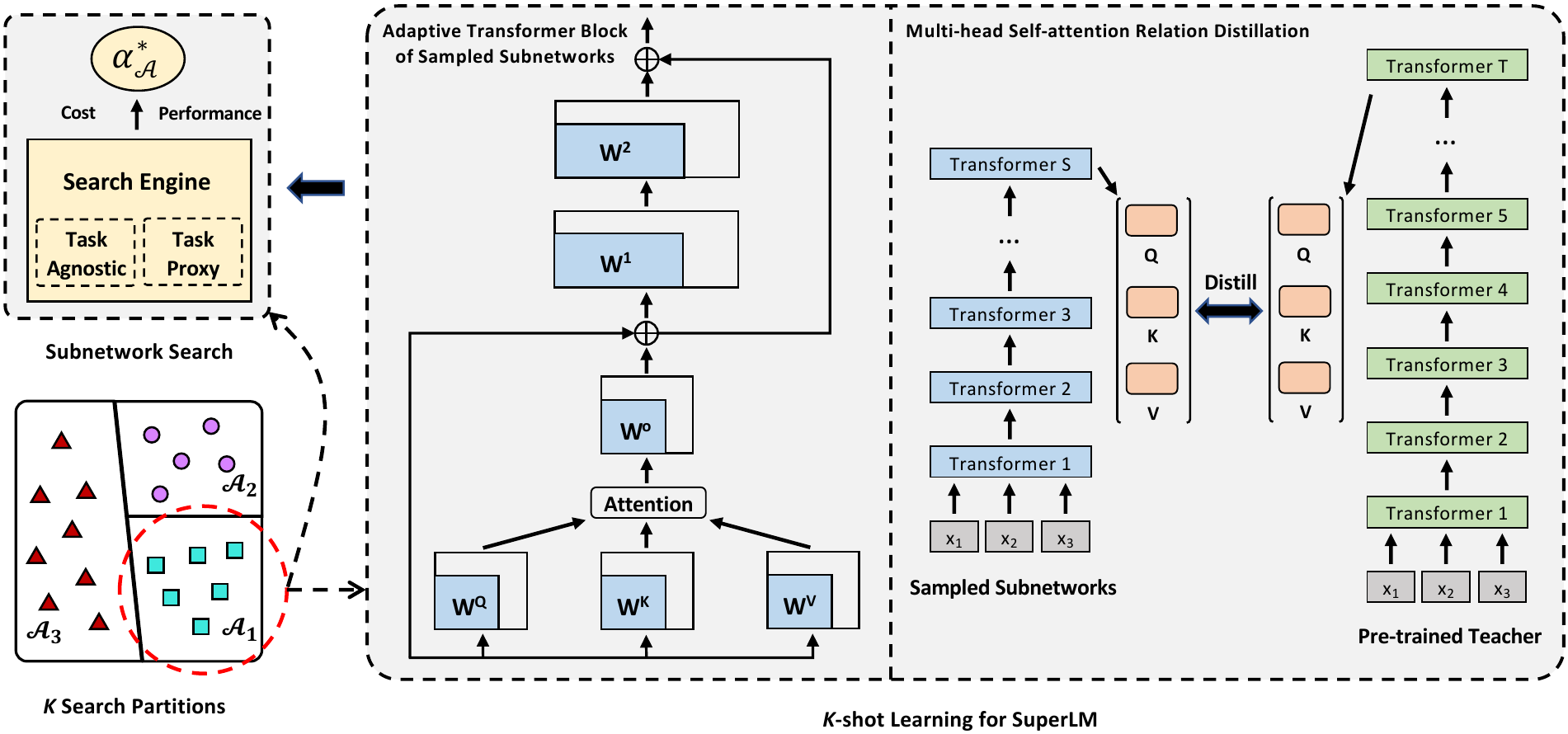}}
\vspace{-0em}
\caption{Overview of \sysname. It considers $K$=$3$ partitions of the Transformer architecture subspace to train one SuperLM for each partition with weight-sharing of the constituent subnetworks trained via task-agnostic deep self-attention distillation. Optimal compressed subnetworks can be easily extracted from the SuperLMs without additional training by task-agnostic or task-proxy search.}
\label{fig:overview}
\end{center}
\vspace{-0.5em}
\end{figure*}

\noindent {\bf Feed-Forward Network (FFN).} Each Transformer layer contains an FNN sub-layer, which is stacked on the MHA. FFN consists of two linear transformations with a ReLU activation as: 
\begin{align}
\label{eq:fnn}
{\rm{FFN}}(x) & = {\rm{max}}(0, x\bm{W}^1 + b_1)\bm{W}^2 + b_2,
\end{align}
where $\bm{W}^1$ $\in$ $\mathbb{R}^{d_{hid}*d_{f}}$, $\bm{W}^2$ $\in$ $\mathbb{R}^{d_{f}*d_{hid}}$, $b_1$ $\in$ $\mathbb{R}^{d_{f}}$, and $b_2$ $\in$ $\mathbb{R}^{d_{hid}}$. In addition, there are residual connection and layer normalization on top of MHA and FFN (denoted by $\oplus$ in Figure~\ref{fig:overview}), which are formulated as {\rm{LayerNorm}}(x + {\rm{MHA}}(x)) and {\rm{LayerNorm}}(x + {\rm{FFN}}(x)), respectively.

\section{Few-shot Task-agnostic NAS}

Given a large pre-trained language model (e.g., BERT) as teacher, \sysname distills several compressed models with variable computational cost in a task-agnostic fashion. In the following, we describe our major components.





\subsection{Search Space Design}
\label{subsec:design}
\begin{table}[!t]

\addtolength{\tabcolsep}{-3.5pt}
\renewcommand{\arraystretch}{1.2}
\centering
\caption{The search space of \sysname with $K$=$3$ partitions, each consisting of $256$ subnets with variable computational cost. We train one SuperLM with weight-sharing for {\em each partition} with child models sharing transformer blocks. Each tuple represents the lowest value, highest value, and steps for each factor.}\vspace{0cm}
\vspace{-0cm}
\label{tb:searchspace}
\begin{tabular}{lccc|c}
\toprule
 & SuperLM$_{\rm{Tiny}}$ &  SuperLM$_{\rm{Small}}$ &  SuperLM$_{\rm{Base}}$  & BERT \\
\midrule
\#Subnets  & 256 & 256 & 256 & N/A \\
\#Layers   & (4, 7, 1)       & (9, 12, 1)      & (9, 12, 1) &  12 \\
\#Hid\_dim & (128, 224, 32)  & (256, 352, 32)  & (544, 640, 32)  & 768 \\
MLP Ratio  & (2.0, 3.5, 0.5) & (2.5, 4.0, 0.5) & (2.5, 4.0, 0.5) &  4.0 \\
\#Heads    & (7, 10, 1)      & (7, 10, 1)      & (9, 12, 1)  &  12 \\
\midrule
\#FLOPs & 40-367$M$ & 0.5-2.1$G$ & 2.1-7.9$G$  & 11.2$G$ \\
\#Params & 4-10$M$ & 12-28$M$  & 39-79$M$  &  109$M$ \\
\bottomrule
\end{tabular}
\vspace{-0.5em}
\end{table}


\noindent {\bf Searchable transformer components.} We presented an overview of Transformers in Section~\ref{sec:background} and our framework in Figure~\ref{fig:overview}. We observe that four important hyper-parameters for the Transformer building blocks, include: 

\begin{itemize}[leftmargin=*]
\topsep=0pt
\partopsep=0pt
\vspace{-0em}
\itemsep=0in
\parsep=0in
\topmargin=0in
\topskip=0in
    \item Number of layers ($L$) to capture the network depth
    \item Hidden dimension ($d_{hid})$ to encode input representation
    \item Attention heads ($H$) for multi-head self-attention
    \item Feed-forward network (FFN) dimension: we encode this by the MLP (multi-layer perceptron) ratio defined as $r=\frac{d_f}{d_{hid}}$ with $d_f$ and $d_{hid}$ representing the intermediate dimension of the FFN and hidden dimension respectively
    \vspace{-0em}

\end{itemize}


All of the above factors are important for model capacity and have a significant impact on the model size and computational cost. For instance, different layers have different feature representation capabilities. Recent works show that Transformer models are overparameterized~\cite{michel2019sixteen,voita2019analyzing}, such as the feed-forward layer (FFN), which is one of the most computation intensive components~\cite{ganesh2020compressing}. Therefore, we search for the optimal MLP ratio and hidden dimension that reduce computational cost resulting from the FFN layers. Furthermore, studies~\cite{NEURIPS2019_2c601ad9,voita-etal-2019-analyzing} show that attention heads can be redundant when they learn to encode similar relationships and nuances for each word. Thus, we make the number of attention heads searchable as well.

\noindent{\bf Inductive bias.} Prior work~\cite{DBLP:journals/corr/RomeroBKCGB14} 
demonstrate that thinner and deeper neural networks with improved representation capacity perform better than wider and shallower ones. We incorporate this as an inductive bias to decide the number of layers to consider for the students in each of our $K$ sub-spaces (base, small, tiny), where we prefer deeper students in terms of the number of layers.  
%
Furthermore, we constrain all the Transformer layers in a given student model to share identical and homogeneous structures, i.e., the same number of attention heads, hidden dimension, etc. This not only reduces the size of the search space, it is also more friendly to hardware and software frameworks~\cite{yin-etal-2021-autotinybert}.


\noindent{\bf Search space partition.} Existing works~\cite{yin-etal-2021-autotinybert,DBLP:conf/kdd/Xu0LS0QL21} 
train a single large SuperLM containing millions of student architectures by weight-sharing. This leads to performance degradation due to optimization interference and convergence of subnetworks with very different sizes~\cite{yu2020bignas}.  
%
To mitigate such interference, we employ a few-shot learning strategy~\cite{chen2021autoformer,zhao2021few} 
as follows: we partition the whole Transformer search space into $K$ sub-spaces such that each sub-space covers different sizes of student models given by the number of parameters. We set $K=3$ to cover typical student sizes, namely base, small and tiny versions. Table~\ref{tb:searchspace} shows the parameter ranges for the $K$ sub-spaces, along with the student configurations contained in each. 

We now encode each sub-space into a SuperLM, where each student model in the space is a subnetwork of the SuperLM. Furthermore, all the student subnetworks share the weights of their common dimensions, with the SuperLM being the largest one in the search space. Considering $K$ independent SuperLMs, each one now has more capacity to encode a sub-space, in contrast to a limited capacity single SuperLM in prior works. Furthermore, our choices for the heuristic partition and inductive bias result in less number of student models of comparable size in each sub-space which alleviates conflicts in weight-sharing.


The student subnetworks are extracted from the SuperLM via bottom-left extraction. In particular, given a specific architecture $\alpha$ = $\{l, d_{hid}, r, h\}$, (i) we first extract alternate $l$ Transformer layers from the SuperLM; (ii) then extract bottom-left sub-matrices in terms of $d_{hid}$ and $r$ from the original matrices that represent the hidden dimension and the MLP ratio respectively; (iii) finally, for the attention heads, we extract the leftmost $h$ heads and retain the dimension of each head as the SuperLM.

\subsection{Task-agnostic SuperLM Training}
We illustrate the SuperLM training process in Algorithm~\ref{alg:training}. Given a large pre-trained language model (e.g., BERT) as the teacher, we initialize the SuperLM with the weights of teacher. In each step of SuperLM training, we randomly sample several student subnetworks from the search space; apply knowledge distillation between the sampled subnetworks and the teacher to accumulate the gradients; and then update the SuperLM. We leverage deep self-attention distillation~\cite{NEURIPS2020_3f5ee243} for task-agnostic training. To this end, we employ multi-head self-attention relation distillation to align the attention distributions as well as the scaled dot-product of keys, queries and values of the teacher and sampled student subnetworks.

Consider $\mathbf{A}_1$, $\mathbf{A}_2$, $\mathbf{A}_3$ to denote the queries, keys and values of multiple relation heads of teacher model, and $\mathbf{B}_1$, $\mathbf{B}_2$, $\mathbf{B}_3$ respectively for a sampled subnetwork. The mean squared error ($\rm{MSE}(\cdot)$) between multi-head self-attention relation of the teacher and sampled subnetwork is used as the distillation objective:
\begin{align}
\mathcal{L} & = \sum_{i=1}^{3}\beta_i \mathcal{L}_i \label{eq:kd-1} \\
\mathcal{L}_i & = \frac{1}{H}\sum_{k=1}^{H}{\rm{MSE}}(\mathbf{R}^T_{ik},\mathbf{R}^S_{ik}) \\
\mathbf{R}^T_{i} = {\rm{softmax}}&(\frac{\mathbf{A}_i\mathbf{A}_i^{\top}}{\sqrt{d_k}}), \ \mathbf{R}^S_{i} = {\rm{softmax}}(\frac{\mathbf{B}_i\mathbf{B}_i^{\top}}{\sqrt{d_k}})
\end{align}
where $H$ is the number of attention heads; $\mathbf{R}^T_{i}$ represents the teacher's $Q-Q$, $K-K$, or $V-V$ relation; $\mathbf{R}^S_{i}$ represents the same for student. $\mathbf{R}^T_{ik}$ is the relation information based on one attention head, and $d_k$ is the attention head size. 

Relation knowledge distillation avoids the introduction of additional parameters to transform the student's representations with different dimensions to align to that of the teacher. 
For the teacher model and subnetworks with different number of attention heads, we first concatenate the self-attention vectors of different attention heads of the subnetwork and then split them according to the number of relation heads of the teacher model. Then, we align their queries with the same number of relation heads for distillation. In addition, we only transfer the self-attention knowledge from the last layer of the teacher model to the last layer of the student model. Automatically selecting which layers to align is an interesting research direction that we defer to future work. 


\begin{algorithm}[tb]
  \caption{Few-shot Task-agnostic Knowledge Distillation with \sysname.}
  \label{alg:training}
\begin{algorithmic}
  \STATE {\bfseries Input:} Partitioned $K$ sub-spaces $\mathcal{A}_k$; initialized $K$ SuperLMs $S_k$ on $\mathcal{A}_k$; pre-trained teacher model $T$; unlabeled data $D$; training epochs $E$; sampling steps $M$
  \STATE {\bfseries Output:} Trained SuperLMs $\{S_k\}$
 \FOR{$k=1$ {\bfseries to} $K$}
  \FOR{$i=1$ {\bfseries to} $E$}
      \STATE Get a batch of data from $D$
      \FOR{$batch$ in $D$}
          \STATE Clear gradients in SuperLM $S_k$
          \FOR{$m=1$ {\bfseries to} $M$}
              \STATE Randomly sample a subnetwork  $s$ from $S_k$
              \STATE Calculate self-attention distil. loss between subnetwork $s$ and teacher $T$ with Eqn.~(\ref{eq:kd-1})
              \STATE Accumulate gradients
          \ENDFOR
          \STATE Update $S_k$ with the accumulated gradients
      \ENDFOR
  \ENDFOR
  \ENDFOR
\end{algorithmic}
\end{algorithm}

Formally, the SuperLM for sub-space $\mathcal{A}_k$ is trained as:
\begin{gather}
\bm{W}^*_{\mathcal{A}_k} = \argmin_{\bm{W}}\mathbb{E}_{\alpha \in \mathcal{A}}[\mathcal{L}(\bm{W}_\alpha; \bm{U}; \mathcal{D}_{train})],
\end{gather}
where, $K$ is the number of sub-space partitions; $\bm{W}$ are the weights of the SuperLM; $\bm{W}_\alpha$ are the weights in $\bm{W}$ specified by the architecture $\alpha$; $\bm{U}$ are the weights of the teacher model including the self-attention module used for distillation; $ \mathcal{D}_{train}$ is the training data set, and $\mathcal{L}(\cdot)$ is the self-attention loss function from Eqn.~(\ref{eq:kd-1}).

\subsection{Lightweight Optimal Student Search}
\label{sec-search-methods}

We outline two search strategies for selecting the optimal student subnetwork.

\noindent{\bf Task-agnostic search.} 
We compute the task-agnostic self-attention distillation loss for all student subnetworks using Eqn.~(\ref{eq:kd-1}) on a heldout validation set from the unlabeled training corpus. The student subnetworks are directly obtained by bottom-left extraction from the well-trained SuperLM (outlined in Section~\ref{subsec:design}). This process is lightweight since it does not require any training or adaptation of the student and the number of subnetworks is limited.

The optimal student is given by the subnetwork 
with the least validation loss subject to the following constraint. 
\begin{gather}
\label{eq:search}
\alpha^*_{\mathcal{A}} = \argmin_{\alpha \in \mathcal{A}_{1, 2, \cdots K}}\mathcal{L}(\bm{W}^*_\alpha; \mathcal{D}_{val}), \quad s.t. \quad g(\alpha) < c,
\end{gather}
where $\bm{W}^*_\alpha$ is the weights of architecture $\alpha$ obtained from $\bm{W}^*_{\mathcal{A}_k}$, $ \mathcal{D}_{val}$ is the validation data set, $\mathcal{L}$ is the self-attention distillation loss, and $g(\cdot)$ is a function to calculate the computational cost (e.g., \#FLOPs, \#parameters) of the subnetwork subject to a given constraint $c$. 

\noindent{\bf Task-proxy search.} This strategy considers a proxy task (e.g., MNLI~\cite{williams-etal-2018-broad}) with label information
to fine-tune each of the $256$ candidate subnetworks in each of the $K$=$3$ sub-spaces. The optimal student in each sub-space is given by the one with the best downstream task performance (e.g., accuracy). Although this strategy is more resource expensive than the task-agnostic one, we demonstrate this to obtain better trade-off in computational cost vs. task performance given the auxiliary task label information.



\section{Experiments}

\subsection{Setup}
\noindent\textbf{Datasets.}
We conduct experiments on the General Language Understanding Evaluation (GLUE) benchmark~\cite{wang2018glue}. We compare our method with the baseline methods on two single-sentence classification tasks (CoLA~\cite{warstadt2018neural}, SST-2~\cite{socher-etal-2013-recursive}), two similarity and paraphrase tasks (MRPC~\cite{dolan2005automatically}, QQP~\cite{chen2018quora}), and three inference tasks (MNLI~\cite{williams-etal-2018-broad}, QNLI~\cite{rajpurkar2016squad}, RTE~\cite{dagan2005pascal,haim2006second,giampiccolo2007third,bentivogli2009fifth})\footnote{We ignore STS-B for a fair comparison with our strongest baseline MiniLM~\cite{NEURIPS2020_3f5ee243} that do not report the task.}. We report accuracy for MNLI, QNLI, QQP, SST-2, RTE, report f1 for MRPC, and report Matthew's correlation for CoLA.

\noindent\textbf{Baselines.}
We compare against several {\em task-agnostic methods}\footnote{For a fair comparison, we do not include DynaBERT~\cite{NEURIPS2020_6f5216f8} with task-specific search, and MobileBERT~\cite{sun2020mobilebert} that uses BERT$_{\rm large}$ as teacher in our main result tables.} generating compressed models from BERT$_{\rm base}$ teacher, using (i) {knowledge distillation} like
BERT$_{\rm{SMALL}}$~\cite{turc2019well}, 
Truncated BERT~\cite{williams-etal-2018-broad}, 
DistilBERT~\cite{sanh2019distilbert}, 
TinyBERT~\cite{jiao2020tinybert},
MINILM~\cite{williams-etal-2018-broad}; as well as those based on Neural Architecture Search, like AutoTinyBERT~\cite{yin-etal-2021-autotinybert}, and NAS-BERT~\cite{DBLP:conf/kdd/Xu0LS0QL21}.
%
%

\noindent\textbf{\sysname configuration.} We use uncased BERT$_{\rm{BASE}}$ as the teacher consisting of $12$ Transformer layers, $12$ attention heads; with the hidden dimension and MLP ratio being 768 and 4, respectively. It consists of $109M$ parameters with $11.2G$ FLOPs. We use English Wikipedia and BookCorpus~ data for SuperLM training with WordPiece tokenization. We use $16$ $V100$ GPUs to train the SuperLM, with $128$ as the batch size and $4e$-$5$ as the peak learning rate for $10$ epochs. The maximum sequence length is set to $128$. The coefficients in distillation objective (Eqn.~(\ref{eq:kd-1})), $\beta_1$, $\beta_2$, and $\beta_3$, are all set to $1$. We distill the self-attention knowledge of the last layer to train the SuperLM. Both the teacher and SuperLM are initialized with pre-trained BERT$_{\rm{BASE}}$.
Other hyper-parameter settings are shown in Appendix.


\begin{table*}[!t]
\small
\addtolength{\tabcolsep}{-2.4pt}
\renewcommand{\arraystretch}{1.2}
\centering
\caption{Performance comparison between models distilled by \sysname against several task-agnostic students ($6$ layer, $768$ hidden size, $12$ heads) distilled from BERT$_{\rm{BASE}}$. We report the relative reduction in computational cost (\#FLOPs and \#Parameters) and improvement in average task performance on GLUE (dev) over all baselines. \sysnameOnlyAuto$_{\rm{Agnostic}}$ is obtained by task-agnostic search. \sysnameOnlyAuto$_{\rm{Proxy_B}}$ and \sysnameOnlyAuto$_{\rm{Proxy_S}}$ are obtained by task-proxy search from SuperLM$_{\rm{base}}$ and SuperLM$_{\rm{small}}$ respectively.} 
\vspace{-0cm}
\label{tb:efficiency}
\begin{tabular}{l|ccc|ccc|ccc}
\toprule 
\multirow{2}*{Model}   & \multicolumn{3}{c}{\sysnameOnlyAuto$_{\rm{Agnostic}}$} & \multicolumn{3}{c}{\sysnameOnlyAuto$_{\rm{Proxy_B}}$} & \multicolumn{3}{c}{\sysnameOnlyAuto$_{\rm{Proxy_{S}}}$} \\
\cmidrule(r){2-4} \cmidrule(r){5-7} \cmidrule(r){8-10}
   & $\Delta$FLOPs & $\Delta$Para & $\Delta$Avg. & $\Delta$FLOPs & $\Delta$Para & $\Delta$Avg. & $\Delta$FLOPs & $\Delta$Para & $\Delta$Avg. \\
\midrule
BERT$_{\rm{BASE}}$~\cite{devlin2019bert} (teacher)  & 81.1\% & 75.5\%  & -2.6 & 60.9\% & 54.3\% & -0.5 & 82.0\% & 76.2\% &  -2.3 \\
BERT$_{\rm{SMALL}}$~\cite{turc2019well}  & 62.4\% & 59.7\% & -0.3 & 22.3\% & 24.7\%  & +1.8  & 64.3\% & 60.8\%  & -0.02 \\
Truncated BERT~\cite{williams-etal-2018-broad}       & 62.4\% & 59.7\% & +2.5  & 22.3\% & 24.7\%  & +4.6  & 64.3\% & 60.8\% & +2.8 \\
DistilBERT\cite{sanh2019distilbert}            & 62.4\% & 59.7\% & +1.1  & 22.3\% & 24.7\%  & +3.2  & 64.3\% & 60.8\% & +1.4 \\
TinyBERT~\cite{jiao2020tinybert}             & 62.4\% & 59.7\% & -0.3 & 22.3\% & 24.7\%  & +1.8  & 64.3\% & 60.8\%  & +0.0 \\
MINILM~\cite{williams-etal-2018-broad}               & 62.4\% & 59.7\% & -1.4 & 22.3\% & 24.7\%  & +0.7  & 64.3\% & 60.8\%  & -1.1 \\
\bottomrule
\end{tabular}
\vspace{-0.2em}
\end{table*}

\begin{table*}[!t]
\small
\addtolength{\tabcolsep}{-3pt}
\renewcommand{\arraystretch}{1.2}
\centering
\caption{Performance comparison between \sysname students, and popular task-agnostic students distilled from BERT$_{\rm{BASE}}$ ($6$ layer, $768$ hidden size, $12$ attention heads). Our results are averaged over 5 runs. Baseline numbers are reported from corresponding papers.}\vspace{0cm}
\vspace{-0cm}
\label{tb:glue}
\begin{tabular}{l|cc|cccccccc}
\toprule
Model & \#FLOPs &  \#Para &  MNLI-m &  QNLI &  QQP &  SST-2 &  CoLA &  MRPC &  RTE &  \multirow{2}*{Average} \\
(Metric) & (G) &  (M) &  (Acc) &  (Acc) &  (Acc) &  (Acc) &  (Mcc) &  (Acc) &  (Acc) & \\
\midrule
BERT$_{\rm{BASE}}$~\cite{devlin2019bert} (teacher)   & 11.2 & 109 & 84.5 & 91.7  & 91.3 & 93.2  & 58.9 & 87.3 & 68.6  & 82.2  \\
BERT$_{\rm{SMALL}}$~\cite{turc2019well}  & 5.66  & 66.5  & 81.8 & 89.8  & 90.6 & 91.2  & 53.5 & 84.9 & 67.9  & 80.0  \\
Truncated BERT~\cite{williams-etal-2018-broad}       & 5.66  & 66.5  & 81.2 & 87.9  & 90.4 & 90.8  & 41.4 & 82.7 & 65.5  & 77.1  \\
DistilBERT\cite{sanh2019distilbert}            & 5.66  & 66.5  & 82.2 & 89.2  & 88.5 & 91.3  & 51.3 & 87.5 & 59.9  & 78.6  \\
TinyBERT~\cite{jiao2020tinybert}             & 5.66  & 66.5  & 83.5 & 90.5  & 90.6 & 91.6  & 42.8 & 88.4 & 72.2  & 79.9  \\
MINILM~\cite{williams-etal-2018-broad}               & 5.66  & 66.5  & 84.0 & 91.0  & 91.0 & 92.0  & 49.2 & 88.4 & 71.5  & 81.0  \\
\midrule
\sysnameOnlyAuto$_{\rm{Agnostic}}$  & 2.13 & 26.8 & 82.8 & 89.9 & 90.8 & 90.6 & 47.1 & 87.3 & 69.0 & 79.6  \\
\sysnameOnlyAuto$_{\rm{Proxy_B}}$   & 4.40 & 50.1 & 83.8 & 90.8 & 91.1 & 91.1 & 55.0 & 88.8 & 71.9 & 81.7  \\
\sysnameOnlyAuto$_{\rm{Proxy_{S}}}$   & 2.02 & 26.1 & 83.2 & 90.0 & 90.6 & 90.1 & 48.3 & 88.3 & 69.4 & 79.9  \\
\sysnameOnlyAuto$_{\rm{Proxy_T}}$   & 0.27 & 6.88 & 79.0 & 86.4 & 89.1 & 85.9 & 24.8 & 78.5 & 64.3 & 72.6 \\
\bottomrule
\end{tabular}
\vspace{-0.2em}
\end{table*}


\begin{table}[!t]
\vspace{-.0em}
\addtolength{\tabcolsep}{-2pt}
\renewcommand{\arraystretch}{1.2}
\centering
\caption{Architecture comparison between the optimal compressed students searched by \sysname with state-of-the-art hand-engineered students distilled from BERT$_{\rm{BASE}}$.}\vspace{0cm}
\vspace{-0cm}
\label{tb:archi}
\begin{tabular}{lcccccc}
\toprule
Model & \#Layers & \#Hid &  Ratio & \#Heads & \#FLOPs & \#Para  \\
\midrule
BERT$_{\rm{BASE}}$ & 12 & 768 & 4 & 12 & 11.2G & 109M \\
MINILM   & 6 & 768 & 4 & 6 & 5.66G & 66.5M \\
\midrule
{\tt AutoDis.$_{\rm{Agnostic}}$}  & 11 & 352 & 4   & 10 & 2.13G & 26.8M \\
{\tt AutoDis.$_{\rm{Proxy_B}}$}   & 12 & 544 & 3   & 9  & 4.40G & 50.1M \\
{\tt AutoDis.$_{\rm{Proxy_{S}}}$} & 11 & 352 & 4   & 8  & 2.02G & 26.1M \\
{\tt AutoDis.$_{\rm{Proxy_T}}$}   & 7  & 160 & 3.5 & 10 & 0.27G & 6.88M \\
\bottomrule
\end{tabular}
\vspace{-0.2em}
\end{table}

\subsection{Finding the Optimal Compressed Models}

We use the following search strategies and constraints to find the optimal compressed models by \sysname.

\noindent {\bf \sysnameOnlyAuto$_{\rm{Agnostic}}$} is obtained by task-agnostic search without any task label information. We set a constraint in Eqn.~(\ref{eq:search}) such that the \#FLOPs of the optimal compressed model is atleast $50\%$ less than the teacher model. We rank all the subnetworks contained in all the partitions of the trained SuperLM by their self-attention distillation loss on the heldout validation set, and select the one that meets the constraint with the minimum loss.

\begin{figure*}[!t]
\vspace{-0cm}
\centering
  \subfigure[ Acc vs \#FLOPs (SuperLM$_{\rm{Base}}$).]{\label{fig:visual-1}
  \includegraphics[width=0.31\textwidth,height=0.22\textwidth]{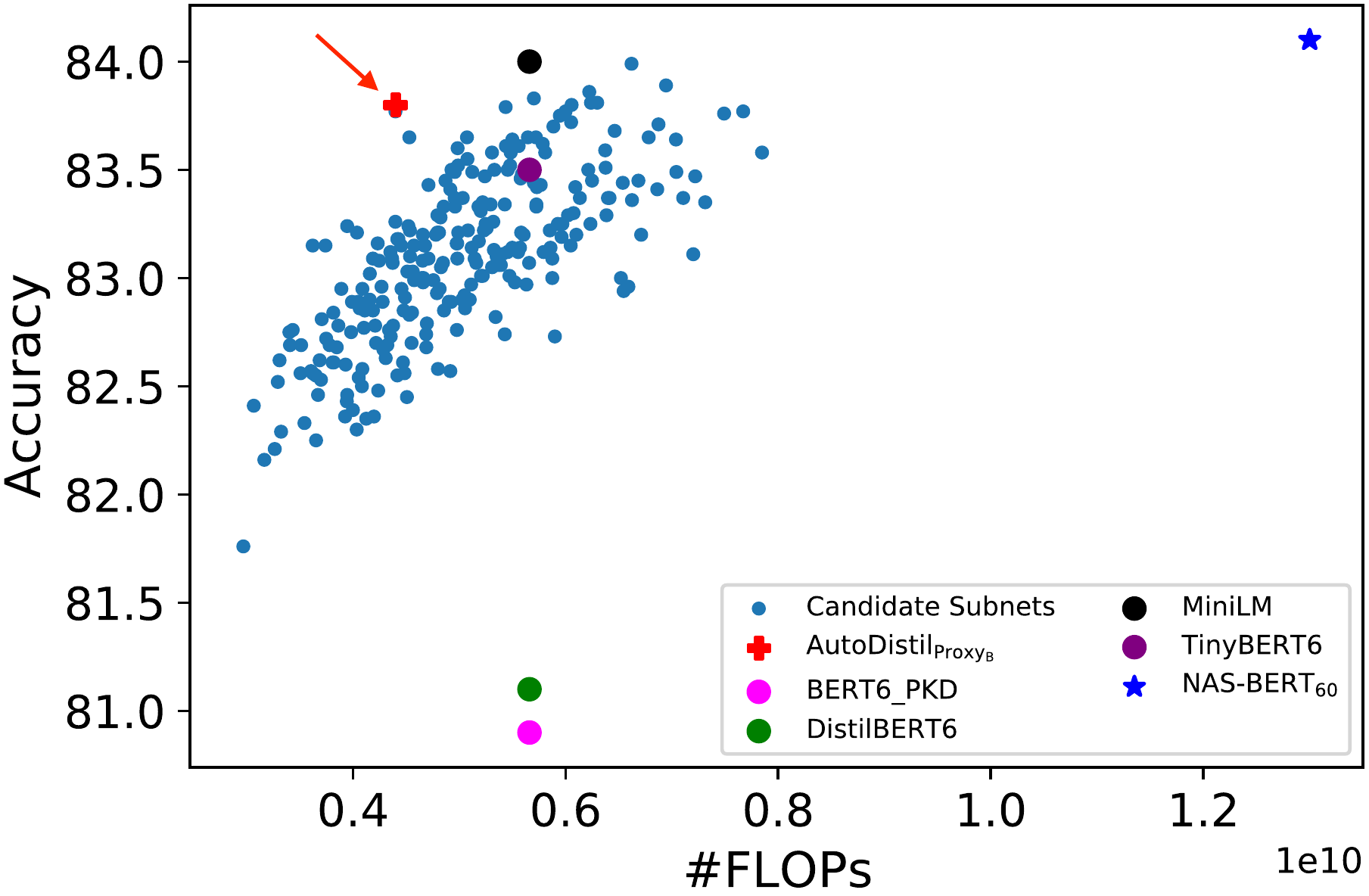}
  }\vspace{0cm}
  \subfigure[Acc vs \#FLOPs (SuperLM$_{\rm{Small}}$).]{\label{fig:visual-2}
  \includegraphics[width=0.31\textwidth,height=0.22\textwidth]{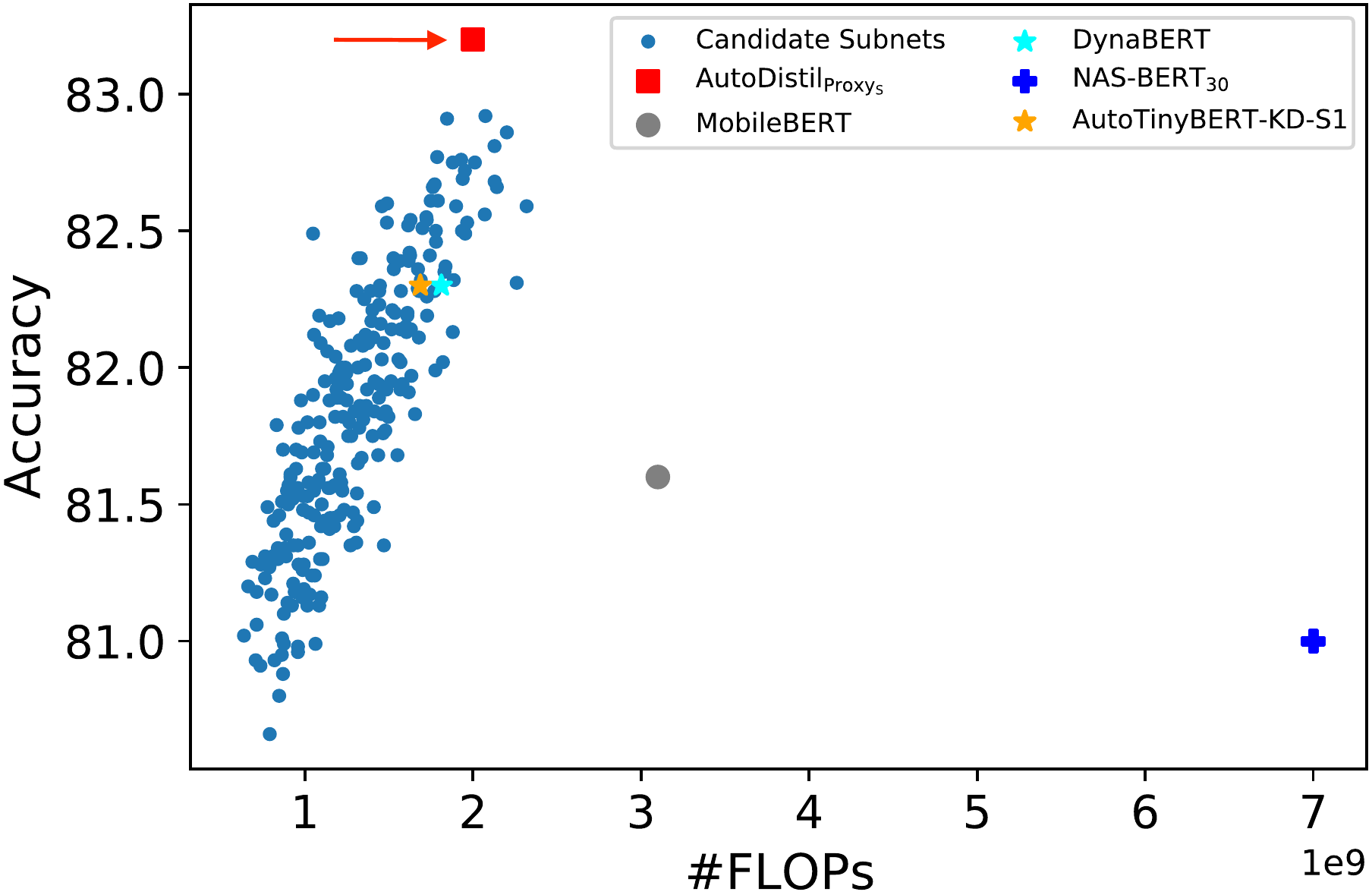}
  }\vspace{0cm}
  \subfigure[Acc vs \#FLOPs (SuperLM$_{\rm{Tiny}}$).]{\label{fig:visual-3}
  \includegraphics[width=0.31\textwidth,height=0.22\textwidth]{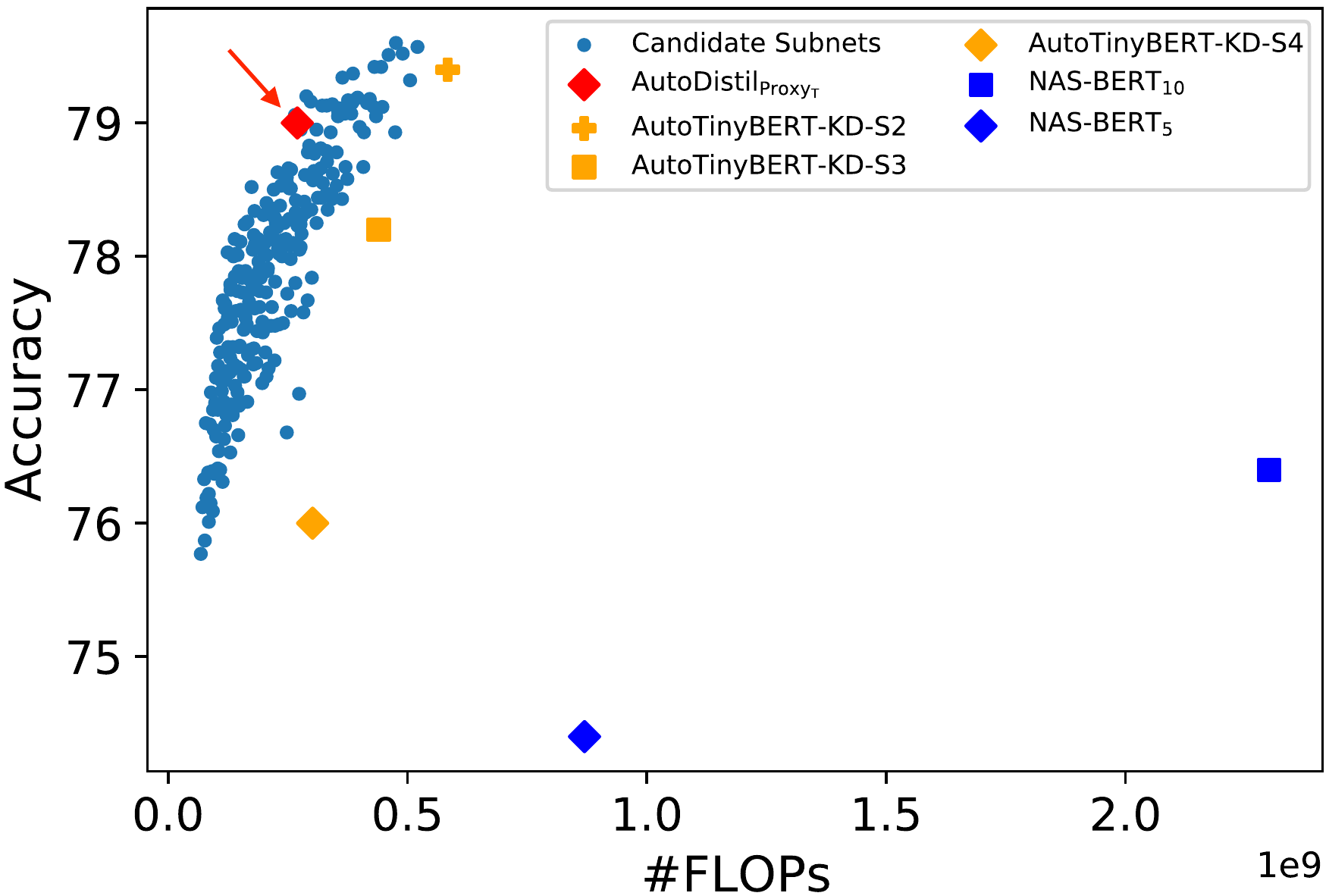}
  }\vspace{-0.0cm}
  \subfigure[Acc vs \#Para (SuperLM$_{\rm{Base}}$).]{\label{fig:visual-4}
  \includegraphics[width=0.31\textwidth,height=0.22\textwidth]{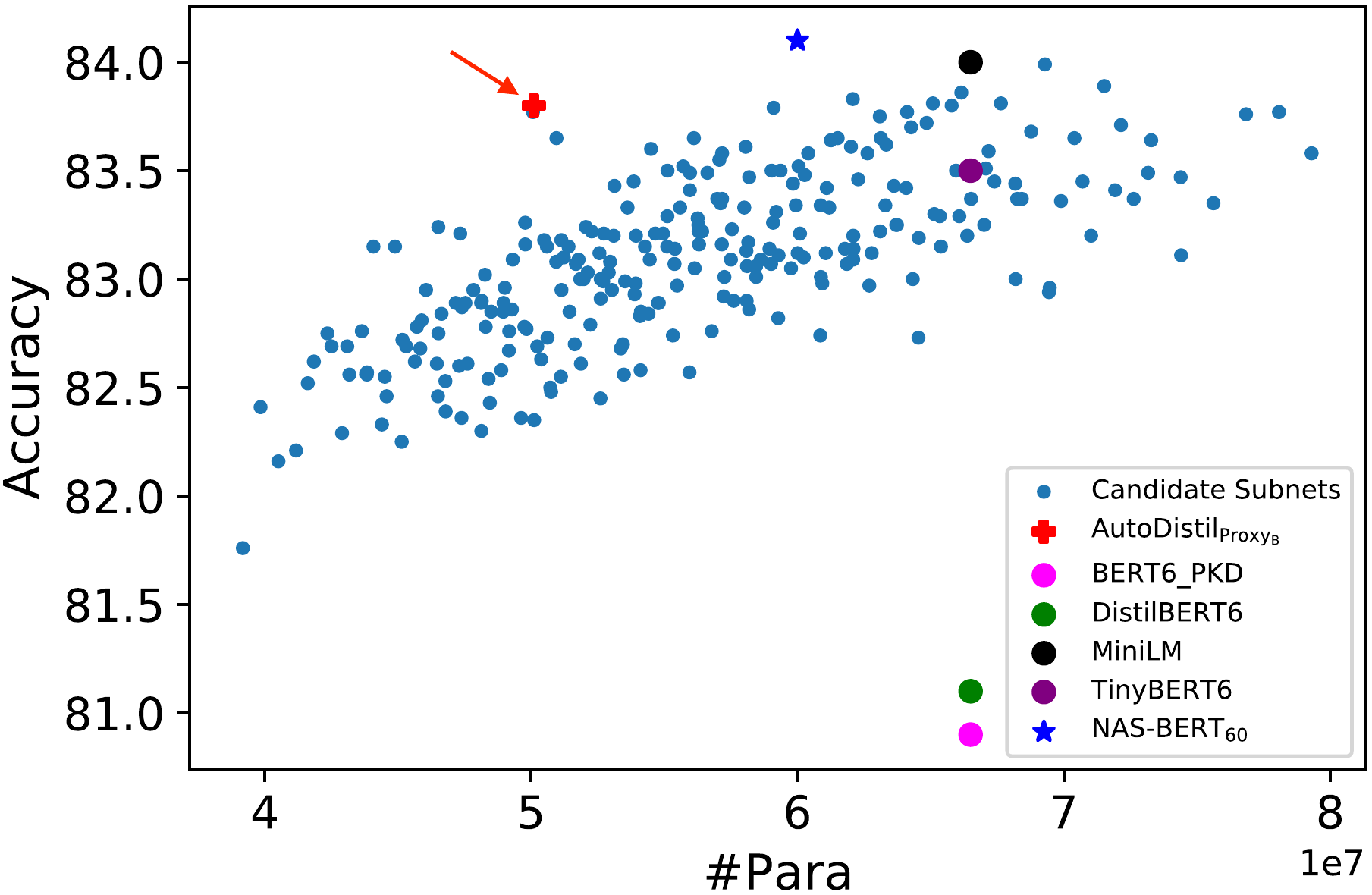}
  }\vspace{0cm}
  \subfigure[Acc vs \#Para (SuperLM$_{\rm{Small}}$).]{\label{fig:visual-5}
  \includegraphics[width=0.31\textwidth,height=0.22\textwidth]{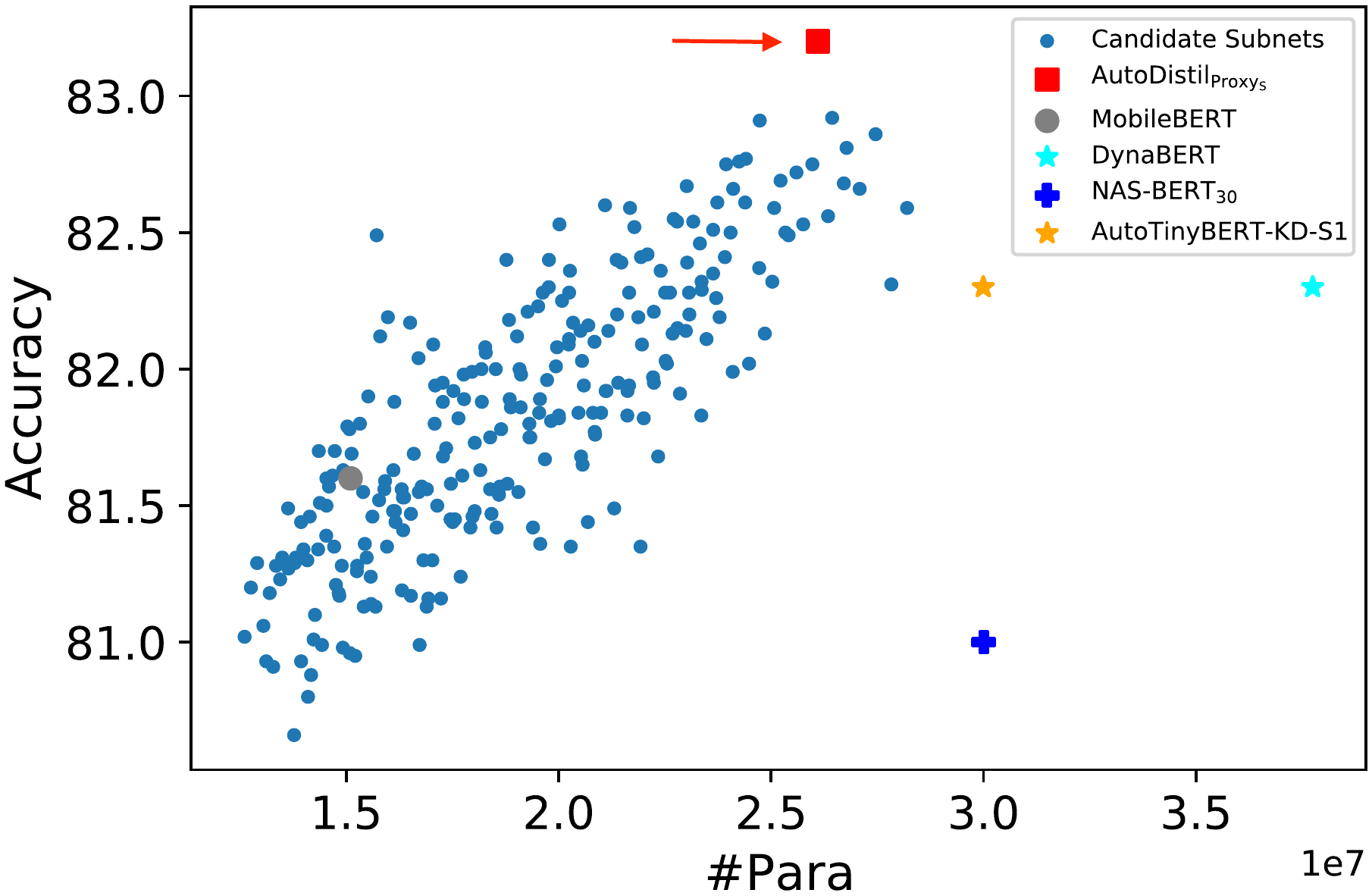}
  }\vspace{0cm}
  \subfigure[Acc vs \#Para (SuperLM$_{\rm{Tiny}}$).]{\label{fig:visual-6}
  \includegraphics[width=0.31\textwidth,height=0.22\textwidth]{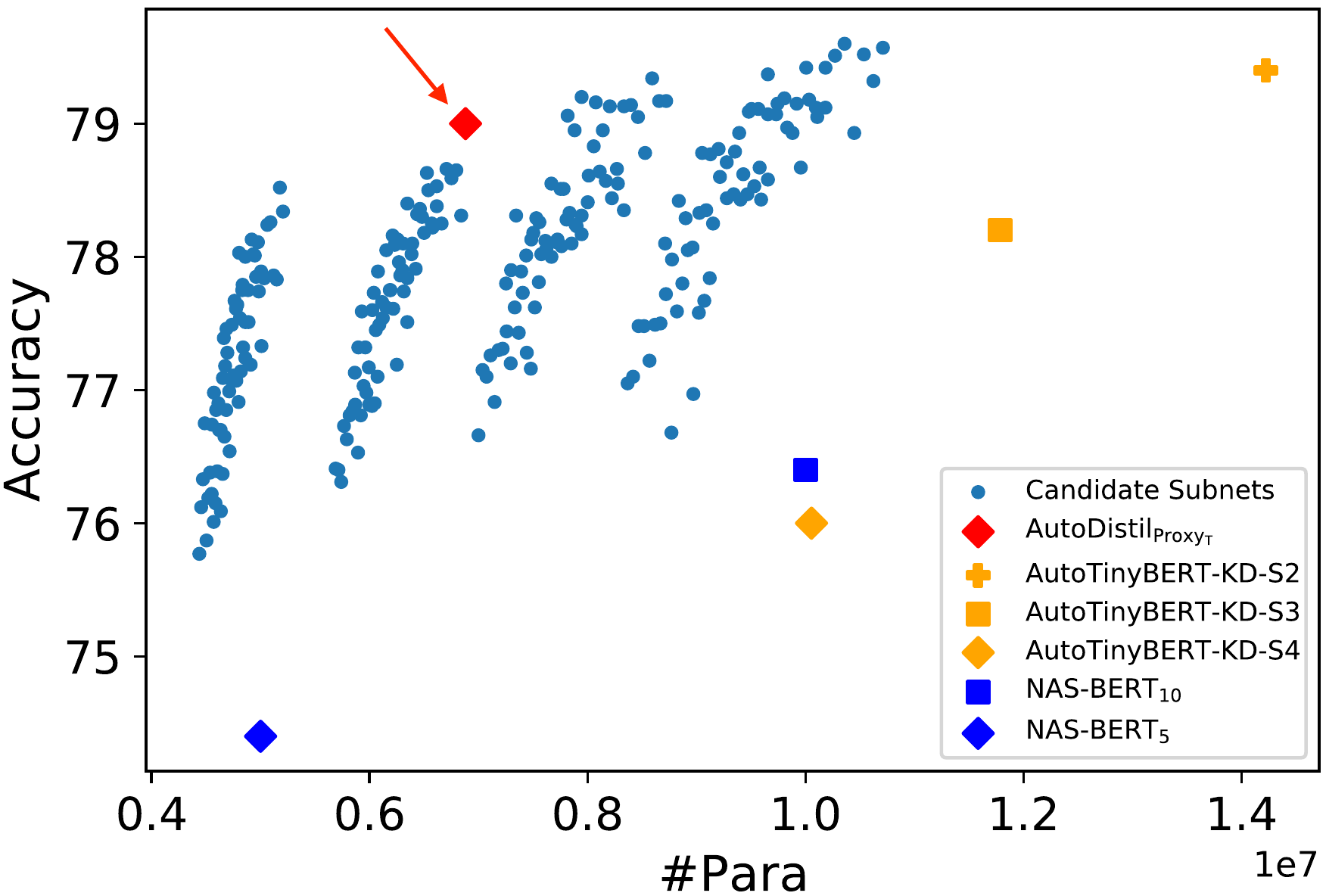}
  }
  \vspace{-0.5em}
\caption{Computational cost vs. task (MNLI) performance trade-off for all $256$ subnetworks contained in each of $K$ SuperLMs (base, small and tiny). \ref{fig:visual-1}-\ref{fig:visual-3} show the trade-off between accuracy (Y-axis) and \#FLOPs (X-axis), and \ref{fig:visual-4}-\ref{fig:visual-6} show the trade-off between accuracy (Y-axis) and \#Para (X-axis). We show the optimal compressed \sysname student for each SuperLM marked in red, along with other state-of-the-art KD and NAS techniques for comparison.} 
\label{fig:visual}\vspace{-0.5em}
\end{figure*}

\noindent {\bf \sysnameOnlyAuto$_{\rm{Proxy}}$} uses MNLI~\cite{williams-etal-2018-broad} as a proxy to estimate downstream task performance of different subnetworks. Prior work~\cite{NEURIPS2020_b6af2c97} has demonstrated performance improvements in MNLI to be correlated to other GLUE tasks. To this end, we fine-tune all subnetworks in each partition of the trained superLMs, and select corresponding subnetworks with the best trade-off between task performance (accuracy) and computational cost (\#FLOPs). This results in $K$=$3$ optimal students, corresponding to \sysnameOnlyAuto$_{\rm{Proxy_{B}}}$, \sysnameOnlyAuto$_{\rm{Proxy_{S}}}$ and \sysnameOnlyAuto$_{\rm{Proxy_{T}}}$ obtained from the corresponding sub-spaces with  SuperLM$_{\rm{Base}}$, SuperLM$_{\rm{Small}}$ and SuperLM$_{\rm{Tiny}}$, respectively. 
We visualize the selected subnetworks for each superLM in Figure~\ref{fig:visual} with the architectures of the optimal compressed models shown in Table~\ref{tb:archi}.

\subsubsection{Comparison with Baselines}
We compare the above \sysname compressed models against state-of-the-art KD and NAS models distilled from the same teacher BERT$_{\rm{BASE}}$. We present the relative performance improvement of \sysname over several baselines in Table~\ref{tb:efficiency} with respect to the following measures: savings in computational cost in the form of (i) FLOPs and (ii) parameter reduction, along with (iii) improvement in the average task performance aggregated over all the GLUE tasks with detailed results in Table~\ref{tb:glue}. 


From Table~\ref{tb:efficiency}, we observe that the compressed model \sysname$_{\rm{Agnostic}}$ generated via our SuperLM training and task-agnostic search has $80\%$ less FLOPs and $75\%$ less parameters, while incurring $2.6$ point accuracy drop in comparison to the large teacher model. When compared to all other baseline models distilled from BERT$_{\rm{BASE}}$, \sysname$_{\rm{Agnostic}}$ leads to $62.4\%$ less FLOPs with $59.7\%$ less parameters while incurring a maximum accuracy drop of less than $1.5$ points -- demonstrating the effectiveness of \sysname in obtaining a better trade-off between task performance and computational cost.

\subsubsection{Search Strategy and Architectures}

From Table~\ref{tb:efficiency}, we observe that both the task-agnostic and task-proxy search strategies achieve better trade-off between performance and cost than the baselines. The compressed model \sysnameOnlyAuto$_{\rm{Proxy_B}}$ obtained from SuperLM$_{\rm{base}}$ by the task-proxy search strategy reduces FLOPs and parameters by $22.3\%$ and $24.7\%$, respectively, while obtaining better task performance than all the baselines. Moreover, by comparing \sysnameOnlyAuto$_{\rm{Agnostic}}$ and \sysnameOnlyAuto$_{\rm{Proxy_S}}$ from SuperLM$_{\rm{small}}$, we observe that task-proxy search obtains better trade-off in cost vs. performance than the task-agnostic one by making use of task label information.

From Table~\ref{tb:archi} we observe that optimal compressed models have thin-and-deep structure which is consistent with findings that thinner and deeper models perform better~\cite{DBLP:journals/corr/RomeroBKCGB14} than wider and shallower ones. While we use this as an inductive bias for sub-space partitioning, our search space (Table~\ref{tb:searchspace}) also contains diverse subnetworks with different depth and width. 
Non-maximal MLP ratio and attention heads for optimal compression indicate that self-attention and feed-forward layers of Transformers are overparameterized~\cite{michel2019sixteen,voita2019analyzing}.

\subsubsection{Subnetwork Performance without Additional Training}
We compare the performance of different student subnetworks generated by \sysname with state-of-the-art NAS and KD techniques in Figure~\ref{fig:visual}. The blue points represent the $256$ subnetworks extracted from each SuperLM and the red points denote the corresponding optimal compressed student, all fine-tuned on the MNLI task. We observe that most of the students (blue points) achieve a good trade-off between performance (accuracy) and cost (\#FLOPs or \#Para) when simply fine-tuned on the downstream task, without additional pre-training or adaptation. Moreover, the optimal compressed students (marked in red) outperform recent NAS methods like NAS-BERT~\cite{DBLP:conf/kdd/Xu0LS0QL21} and AutoTinyBERT~\cite{yin-etal-2021-autotinybert} that perform an additional stage of pre-training or distillation of the candidate students obtained from NAS. 
More than half of the subnetworks in Figure~\ref{fig:visual-4} also show better trade-off than the best task-agnostic KD method MiniLM~\cite{NEURIPS2020_3f5ee243}. These observations demonstrate the effectiveness of our few-shot task-agnostic SuperLM training and search mechanism.


\subsubsection{Task-agnostic Training Strategies}
\label{exp-kd-strategy}
We study different task-agnostic strategies for SuperLM training in \sysname. 
Specifically, we compare three strategies in Table~\ref{tb:effect-kd}. 
(i) Replacing the KD loss in Eqn.~(\ref{eq:kd-1}) with masked language modeling (MLM) loss~\cite{devlin2019bert} to calculate gradients. (ii) KD$_{att}$+Cont further continues training the searched compressed models on the large language corpus. (iii) KD$_{att}$ is the strategy adopted in \sysname for self-attention distillation.
We evaluate subnetworks with the same architecture ($6$ layers, $768$ hidden, $12$ heads, MLP ratio $4$) from the trained SuperLM. We fine-tune the subnetworks on RTE and MRPC tasks, and report accuracy and f1 respectively. First, we observe self-attention distillation to perform better than MLM, for SuperLM training. Second, we observe limited performance gains with continued training demonstrating the effectiveness of our single-stage training protocol.

\begin{table}
\addtolength{\tabcolsep}{8.5pt}
\renewcommand{\arraystretch}{1}
\centering
\caption{Comparing task-agnostic SuperLM training strategies.}\vspace{0cm}
\vspace{-0cm}
\label{tb:effect-kd}
\begin{tabular}{lcc}
\toprule
Strategy &  MRPC &  RTE\\
\midrule 
MLM   & 89.4  & 68.2  \\
KD$_{att}$+Cont.   & 91.0  & 71.8\\
KD$_{att}$   & 91.2  & 71.5 \\
\bottomrule
\end{tabular}
\vspace{-0.2em}
\end{table}

\begin{table}
\addtolength{\tabcolsep}{5.0pt}
\renewcommand{\arraystretch}{1}
\centering
\caption{Comparing search space design strategies.}\vspace{0cm}
\vspace{-0cm}
\label{tb:effect-space}
\begin{tabular}{l|ccc|c}
\toprule
\multirow{2}*{Task}  & \multicolumn{4}{c}{Search Space Size (number of subnetworks)}  \\
&\multicolumn{3}{c|}{One-shot} & $K$=$3$-shot\\
\cmidrule(r){2-5}
& 27 &  864 &  11232 &  256*3  \\
\midrule
MRPC   & 88.2  & 87.5  & 85.1 & 91.2 \\
RTE    & 67.2  & 64.5  & 62.8 & 71.8 \\
\bottomrule
\end{tabular}
\vspace{-0.2em}
\end{table}

\subsubsection{Search Space Design Strategies}
In Table~\ref{tb:effect-space}, we compare one-shot NAS versus few-shot NAS training for our SuperLM. For one-shot NAS, we consider a single search space containing different numbers of subnetworks (e.g., $27$, $864$, $11232$). For few-shot NAS, we consider $K$=$3$ sub-spaces containing $256$ subnetworks each. We extract subnetworks with the same architecture ($6$ layers, $768$ hidden, $12$ heads, MLP ratio $4$) from trained SuperLM for each strategy for evaluation. We fine-tune the subnetworks on RTE and MRPC tasks, and report accuracy and f1 respectively. We observe fewer subnetworks contained in a single search space for one-shot NAS result in a better performance. This results from optimization interference as the number and size of subnetworks increase. Finally, we observe our design strategy with few-shot NAS to perform the best while containing lesser number of subnetworks. 

\section{Related Work}

\noindent {\bf Task-specific knowledge distillation.} Knowledge distillation (KD)~\cite{44873} is one of the most widely used techniques for model compression, which transfers knowledge from a large teacher to a smaller student model. 
Task-specific distillation aims to generate smaller student models by using downstream task label information. Typical task-specific distillation works include  
BERT-PKD~\cite{sun2019patient}, 
BERT$_{\rm{SMALL}}$~\cite{turc2019well}, 
TinyBERT~\cite{jiao2020tinybert}, 
DynaBERT~\cite{NEURIPS2020_6f5216f8}, and SparseBERT~\cite{xu2021rethinking}. 
While task-specific KD methods often achieve good task performance, a typical drawback is that it is resource-consuming to run distillation for each and every task, and also not scalable. 

\noindent{\bf Task-agnostic knowledge distillation.} In contrast to task-specific distillation, we explore task-agnostic KD that does not use any task label information. The distilled task-agnostic models can be re-used by simply fine-tuning on downstream tasks. They can also be used to initialize students for task-specific distillation. Task-agnostic distillation leverages knowledge from soft target probabilities, hidden states, layer mappings and self-attention distributions of teacher to train student models. Typical task-agnostic distillation works include 
DistilBERT~\cite{sanh2019distilbert}
MobileBERT~\cite{sun2020mobilebert},
and MiniLM~\cite{NEURIPS2020_3f5ee243}. 
MobileBERT assumes that students have the same number of layers as the teacher for layer-by-layer distillation. MiniLM transfers self-attention knowledge from the last layer of the teacher to that of the student. These works rely on hand-designed architecture for the student models for KD that requires several trials, and needs to be repeated for a new student with a different cost. In contrast, we develop techniques to automatically design and distill several student models with variable cost using NAS.

\noindent{\bf Neural Architecture Search.} While NAS has been extensively studied in computer vision~\cite{pham2018efficient,tan2019mnasnet,cai2020once,yu2020bignas}, there has been relatively less exploration in natural language processing. 
Evolved Transformer~\cite{pmlr-v97-so19a} and HAT~\cite{hanruiwang2020hat} search for efficient sub-networks from the Transformer architecture for machine translation tasks.
Some recent approaches closest to our method include, DynaBERT~\cite{NEURIPS2020_6f5216f8}, AutoTinyBERT~\cite{yin-etal-2021-autotinybert} and NAS-BERT~\cite{DBLP:conf/kdd/Xu0LS0QL21}. 
While DynaBERT performs task-specific distillation, AutoTinyBERT uses task-agnostic KD and MLM strategies for SuperLM training, but task-specific search for the compressed models.
NAS-BERT uses a different search space, and performs two-stage knowledge distillation with pre-training and fine-tuning of the candidates. Both of these approaches employ one-shot NAS as a single large search space containing millions of subnetworks that result in co-adaption and weight-sharing challenges between them for SuperLM training. In contrast, our method employs few-shot NAS with a compact search space design to address the above challenges. This further allows us to do a lightweight search for the optimal student without re-training in a fully task-agnostic fashion.

\section{Conclusion}
We develop a few-shot task-agnostic NAS framework, namely \sysname for distilling large language models into compressed students with variable computational cost. 
To address the co-adaption and weight-sharing challenges for SuperLM training, we partition the Transformer search space into $K$=$3$ compact sub-spaces covering important architectural components like the network depth, width, and number of attention heads. We leverage deep self-attention distillation for fully task-agnostic SuperLM training and lightweight optimal student search without re-training. This allows our students to be re-used by simply fine-tuning on downstream tasks.
Experiments in the GLUE benchmark demonstrate that \sysname outperforms state-of-the-art task-agnostic distillation methods with $62.4\%$ less computational cost and $59.7\%$ less parameters while obtaining a similar downstream task performance.


\bibliographystyle{unsrtnat}
\bibliography{main}  






\newpage
\appendix
\onecolumn
\section{Appendix}

\subsubsection{Comparison with Baselines}

\begin{figure}[!ht]
\begin{center}
\centerline{\includegraphics[width=0.48\textwidth,height=0.32\textwidth]{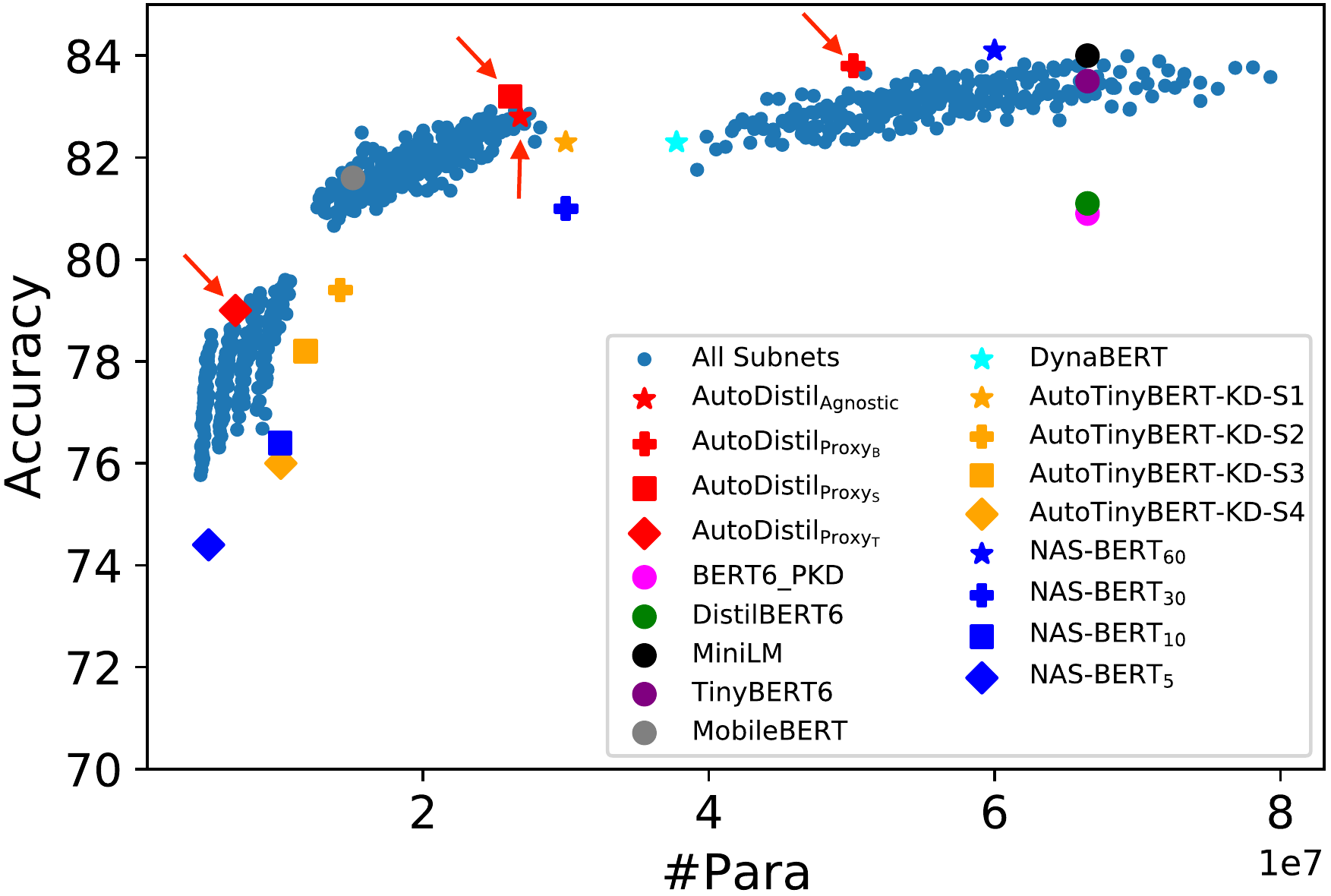}}
\vspace{-0em}
\caption{Comparison between \sysname and state-of-the-art distilled models.}
\label{fig:acc_para}
\end{center}
\vspace{-0.0em}
\end{figure}

We compare \sysname with state-of-the-art distilled models in terms of the trade-off between model size (\#Para) and performance (accuracy). The results are shown in Figure~\ref{fig:acc_para}.
\sysname uses few-shot task-agnostic Neural Architecture Search to distill several compressed students with variable \#Para (x-axis) from $K$=$3$ SuperLMs (corresponding to each point cloud) trained on $K$ sub-spaces of Transformer search space. Each student extracted from the SuperLM is fine-tuned on MNLI with y-axis showing accuracy. The best student from each SuperLM is marked in red. Given any state-of-the-art distilled model, \sysname generates a better candidate with less \#Para and improved task performance from corresponding search space.

\subsubsection{Layer Selection Strategies}

\begin{table}[!h]
\addtolength{\tabcolsep}{6.0pt}
\renewcommand{\arraystretch}{1.2}
\centering
\caption{Effects of layer selection strategies.}\vspace{0cm}
\vspace{-0cm}
\label{tb:effect-layer}
\begin{tabular}{l|c|c}
\toprule
Strategy  & MRPC &  RTE  \\
\midrule
Alternate\_Dropping & 91.2  & 71.8 \\
Top\_Dropping & 90.6  & 68.5  \\
Alternate\_Top\_Dropping    & 85.7  & 62.7 \\
\bottomrule
\end{tabular}
\vspace{-0.0cm}
\end{table}

We study different strategies to construct subnetwork layers by selecting layers from the superLM model. Alternate\_Dropping is the strategy adopted in \sysname such that we drop alternating odd layers from the superLM model to construct subnetwork layers. Top\_Dropping means that we drop top layers of superLM to construct subnetwork layers. Alternate\_Top\_Dropping means that we first perform Alternate\_Dropping in superLM training stage and then perform Top\_Dropping in fine-tuning stage (please refer to~\cite{sajjad2020effect} for more details of different layer selection strategies).
For all strategies, we perform knowledge distillation between the last layer of the teacher model and the last layer of the subnetworks.
We evaluate the subnetworks with the same architecture (\#layer=6, \#hid=768, R=4, \#heads=12) after superLM is trained. We report accuracy and f1 for RTE and MRPC, respectively.

We report the results in Table~\ref{tb:effect-layer}. We observe that the strategy of Alternate\_Dropping achieves the best performance on both MRPC and RTE tasks, which demonstrates the effectiveness of the layer selection strategy used in \sysname. Alternate\_Top\_Dropping performs the worst due to interference when different layer selection strategies are used in the superLM training stage and the fine-tuning stage of compressed models. This indicates that the knowledge contained in the superLM model and the compressed model is structured and that it is non-trivial to select layers from superLM to extract subnetwork layers.

\subsubsection{Scaling of Training Data}

\begin{table}[!h]
\addtolength{\tabcolsep}{2.0pt}
\renewcommand{\arraystretch}{1.2}
\centering
\caption{Scaling of training data.}\vspace{0cm}
\vspace{-0cm}
\label{tb:effect-data}
\begin{tabular}{l|c|c|c|c}
\toprule
\multirow{2}*{Strategy}  & MNLI  &  ParaNMT &  Wiki & Wiki+Book  \\
  & (393k)  &  (5M) &  (29M) & (40M) \\
\midrule
MRPC   & 88.3 & 88.2  & 89.4 & 91.2 \\
RTE    & 65.4 & 67.2  & 68.6 & 71.8 \\
\bottomrule
\end{tabular}
\vspace{-0.0cm}
\end{table}

We investigate the effects of data sets of different sizes used for superLM training. In particular, we compare MNLI~\cite{williams-etal-2018-broad}, ParaNMT~\cite{wieting-gimpel-2018-paranmt} (we sampled 5 million samples from the original 50 million data), Wiki, and Wiki+Book~\cite{zhu2015aligning}. We report the size of each data set and the performance of \sysname with each training data set in Table~\ref{tb:effect-data}. 
We observe that \sysname performs the best with Wiki+Book data set, and the larger the data set, the better the performance. Moreover, we observe similar performance for MNLI and ParaNMT data sets, especially on MRPC task. This is because MNLI is correlated to other GLUE tasks. In addition, we observe that an increase in the amount of data does not guarantee to bring an equivalent increase in performance. For example,  Wiki data set is more than five times larger than ParaNMT data set, but our method performs only about 1\% better With Wiki data set than with ParaNMT. These observations illustrate that while using a larger data set does improve the performance of the method, the improvement could be quite limited. 


\subsubsection{Hyper-parameter Settings for Fine-Tuning}

\begin{table}[!h]
\addtolength{\tabcolsep}{1pt}
\renewcommand{\arraystretch}{1.2}
\centering
\caption{Hyper-parameters used for fine-tuning on GLUE.}\vspace{0cm}
\vspace{-0cm}
\label{tb:hyper-para}
\begin{tabular}{l|c|c|c}
\toprule
Tasks  & Learning Rate & Batch Size & Epochs \\
\midrule
MNLI-m & 2e-5  & 32  & 5 \\
QNLI   & 2e-5  & 32  & 5 \\
QQP    & 2e-5  & 32  & 5 \\
SST-2  & 2e-5  & 32  & 10 \\
CoLA   & 1e-5  & 32  & 20 \\
MRPC   & 2e-5  & 32  & 10 \\
RTE    & 2e-5  & 32  & 10 \\
\bottomrule
\end{tabular}
\vspace{-0.0cm}
\end{table}

We report the fine-tuning hyper-parameter settings of GLUE benchmark in Table~\ref{tb:hyper-para}. \sysname and baselines follow the same settings.

\end{document}